\documentclass[conference]{IEEEtran}
\usepackage[utf8]{inputenc}
\usepackage{times}

\newcommand{\bic}{Ball-in-a-Cup}

\title{Simultaneously Learning Vision and Feature-based Control Policies for Real-world \bic{}}

\author{\authorblockN{Devin Schwab$^{\dagger *}$ \and Tobias Springenberg$^\ddagger$ \and Murilo F. Martins$^\ddagger$ \and Thomas Lampe$^\ddagger$ \and Michael Neunert$^\ddagger$ \and Abbas Abdolmaleki$^\ddagger$ \and Tim Hertweck$^\ddagger$ \and Roland Hafner$^\ddagger$ \and Francesco Nori$^\ddagger$ \and Martin Riedmiller$^\ddagger$}\and
\authorblockN{$^\dagger$ Robotics Institute, Carnegie Mellon University, Pittsburgh, Pennsylvania}\and
\authorblockN{$^\ddagger$ Google DeepMind, London, GB}}

\usepackage[numbers]{natbib}
\usepackage{graphicx}
\usepackage[normalem]{ulem}
\usepackage[bookmarks=true]{hyperref}
\usepackage{amssymb}
\usepackage{amsmath}
\usepackage[mode=buildnew]{standalone}
\usepackage[mathscr]{euscript}
\usepackage{multirow}
\usepackage{booktabs}

\newcommand{\cM}{{\mathcal{M}}}
\newcommand{\cA}{{\mathcal{A}}}
\newcommand{\cT}{{\mathcal{T}}}
\newcommand{\sA}{\mathscr{A}}
\newcommand{\sS}{\mathscr{S}}
\newcommand{\cB}{{\mathcal{B}}}

\newcommand\blfootnote[1]{%
    \begin{NoHyper}
    \renewcommand\thefootnote{}\footnote{#1}%
    \addtocounter{footnote}{-1}%
    \end{NoHyper}
}

\allowdisplaybreaks

\begin{document}

\maketitle

\begin{abstract}
    We present a method for fast training of vision based control policies on real robots. The key idea behind our method is to perform multi-task Reinforcement Learning with auxiliary tasks that differ not only in the reward to be optimized but also in the state-space in which they operate. In particular, we allow auxiliary task policies to utilize task features that are available only at training-time. This allows for fast learning of auxiliary policies, which subsequently generate good data for training the main, vision-based control policies. This method can be seen as an extension of the Scheduled Auxiliary Control (SAC-X) framework. We demonstrate the efficacy of our method by using both a simulated and real-world \bic{} game controlled by a robot arm. In simulation, our approach leads to significant learning speed-ups when compared to standard SAC-X. On the real robot we show that the task can be learned from-scratch, i.e., with no transfer from simulation and no imitation learning.
    Videos of our learned  policies running on the real robot can be found at https://sites.google.com/view/rss-2019-sawyer-bic/.
\end{abstract}

\section{Introduction}

\blfootnote{$^*$Correspondence to dschwab@andrew.cmu.edu}

Automated learning of robot control from high-dimensional sensor data -- such as camera images -- is a long standing goal of robotics research. In recent years, a lot of progress towards this challenge has been made by employing machine learning methods for real robot learning. Examples of such successes include: learning grasp success predictors from raw image data~\citep{levine2018learning} and the growing list of successful real-world applications of Reinforcement Learning (RL) such as learning to flip a pancake~\cite{kormushev2010robot}, learning to fly a model helicopter~\cite{abbeel2007application}, in-hand manipulation of objects using a Shadow Robotics hand~\cite{andrychowicz2018learning} and the \bic{} task with a robot arm~\cite{kober2009learning}. However, most of these approaches have either used extensive pre-training in simulation, used expert demonstrations to speed-up learning, or did not consider learning from raw sensory input (e.g. images). 
Most of these limitations are due to the fact that only a limited amount of data is available when learning from a real robot system, prohibiting the naive application of expressive control policies (such as deep neural networks learned from vision).

One common strategy for minimizing the need for large amounts of data is to carefully curate the information used to represent the state of the robotic system (from which an optimal action should be predicted). Defining the state of the robot via expressive features -- e.g. positions, orientations and velocities of the robotic system and considered objects in a scene -- can drastically simplify the learning problem since: i) the learned control policies require much fewer parameters to represent a solution, and ii) the learned policies do not need to learn to recognize objects of interest from images (and conversely learn to explicitly ignore spurious signals such as distractors in the visual scene).

While the above makes a compelling case for using feature based representations when learning in the real world, feature-extraction incurs a significant overhead in the experimental setup; as computing meaningful features from real-world sensor data can be difficult, computationally expensive and requires extensive sensor arrays and pre-processing pipelines.
Furthermore, even when such a feature extraction pipeline is available during learning, it is often desirable to obtain a controller that -- once learned successfully -- can be deployed with a minimal setup (e.g. in our experiments below an object tracking system is required during training but not afterwards).

In this paper, we develop a method that can simultaneously learn control policies from both feature-based representation and raw vision input in the real-world -- resulting in controllers that can afterwards be deployed on a real robot using two off-the-shelf cameras. To achieve this we extend the multi-task RL framework of Scheduled Auxiliary Control (SAC-X) \citep{riedmiller2018learning} to scenarios where the state-space for different tasks is described by different modalities. We base our approach on the following assumptions:

\begin{enumerate}
    \item An extended set of observations is available to the agent during training time, which includes raw sensor data (such as images), proprioceptive features (such as joint angles), and ``expensive'' auxiliary features given via an object tracking system.
    \item At test time, the ``expensive'' auxiliary features are unavailable, and the policy must rely only on raw sensors data and proprioceptive information.
    \item Control policies learned with features will converge faster than policies learned directly from raw-sensor data, as shown in Figure~\ref{fig:simulation-diff-state-vs-baseline}. And hence provide a useful guiding signal for learning vision based policies.
    \item A subset of the observations provided to the learning system forms a state description for the MDP of the RL problem we want to tackle (i.e. given this subset and a control action it is possible to predict the next state).
\end{enumerate}

Starting from these assumptions we develop a method with the following properties i) like SAC-X our approach allows us to jointly learn a control policy for one main task and multiple auxiliary tasks, ii) our method allows for joint learning of vision and feature based policies for all tasks, iii) through a special neural-network architecture, all learned policies internally share a set of features (i.e. they share hidden layers in the network) enabling accelerated learning, and iv) by executing feature based skills during training the convergence of vision based control policies can be improved.

While the presented approach is general, we focus on one concrete application: solving the \bic{} task using a robotic arm. This task requires a robot arm to swing-up a ball attached to a string and catch it in a cup (c.f. Figures \ref{fig:real-robot-setup} and \ref{fig:cup-components} for a visualization). We show, in simulation, that our method significantly accelerates learning when compared to learning from raw image data without auxiliary features like object positions. Furthermore, we demonstrate that our method is capable of learning the task on real hardware. Compared to previous solutions to the \bic{} task~\citep{kober2009learning}, our method does neither require imitation data nor a specialized policy class and can learn the task in approximately 28 hours of {training}\footnote{Training time ignores the time taken for the robot to automatically reset the environment. Observed overhead of 20-30\% due to resets and entanglement.} on a single robot.

\section{Related Work}
Our method can be seen as an extension of a long line of work on aiding exploration in Reinforcement Learning with auxiliary tasks (defined via different reward functions). Among these is the idea of General value functions, as coined in work on learning with pseudo-rewards associated with random goals defined based on sensorimotor datastreams ~\cite{sutton2011horde}. This idea was further extended to a Deep RL setting in work on Universal Value Function Approximators (UVFA)~\cite{schaul2015universal}, which aims to learn a goal conditioned action-value function. In a further extension, Hindsight Experience Replay (HER) ~\cite{andrychowicz2017hindsight} introduced the idea of computing rewards for different goals ``in hindsight'' by treating the end of a recorded trajectory as the intended goal. Finally, Scheduled Auxiliary Control (SAC-X)~\cite{riedmiller2018learning} -- the method on which we base our paper -- aims to simultaneously learn so called intention-policies for several, semantically grounded, auxiliary tasks and use each intention to drive exploration for all other tasks. 

Just like SAC-X, our approach can also be related to curriculum learning approaches (see e.g.~\citet{bengio2009curriculum} for a description and \citet{heess2017emergence} for an RL application). In particular, our experiments utilize multiple reward functions with varying amounts of shaping and difficulty of success. While many curriculum learning approaches invent new tasks based on internal metrics such as current task performance~\citep{schmidhuber2013powerplay,forestier2017intrinsically,jiang2018mentornet} we here restrict ourselves to a fixed set of pre-specified tasks. These algorithms also often employ methods for intelligently scheduling tasks to learn based on learning progress. We here use simple uniform random selection of tasks. Each task is executed for a fixed number of steps before a new task is chosen.

While numerous works exist that consider RL in multi-task settings \citep{hessel2018popart,parisotto2015actor,espeholt2018impala} these typically focus on varying rewards and environment behaviors but expect a fully shared observation space. Work on learning with varying observations across tasks is more scarce (we refer to  \citet{espeholt2018impala} for a recent success). Among these the asymmetric actor-critic technique~\citep{pinto2017asymmetric,andrychowicz2018learning} is closest to our approach. There the authors experiment with the idea of providing additional state-information (that is only available at training time) to a learned action-value function (while withholding said information from the policy) to ease credit assignment. Our approach generalizes this idea to using a different definition of state not only for the actor and critic, but also across different tasks -- while still learning a single policy network. We further expand on this interesting relation in Section \ref{sec:asym-actor-critic-diff-state-spaces}.

Our work also has loose connections to ideas from behavioral cloning~\citep{abbeel2004apprenticeship,ross2011reduction}, a simple form of imitation learning where an agent attempts to replicate another agent's policy. Since we duplicate tasks for feature-based and vision-based policies (i.e. the tasks obtain the same reward and differ only in their state-spaces), the feature-based policies can be seen as generating demonstration data for the vision based policies (or vice versa) in our experiments. While no direct behavior cloning is performed, transitions generated from one policy are used for off-policy learning of other policies.

In this paper we test and demonstrate our technique on the dynamic \bic{} task. Robot learning frameworks that utilize RL have been applied to many different robots and many different tasks, including recent successes such as picking up simple objects~\citep{rusu2016sim}, manipulating objects in hand~\citep{andrychowicz2018learning}, and manipulating a valve~\cite{haarnoja2018softb}. However, many of these recent works focus on mostly static tasks, such as object manipulation. Dynamic tasks have additional difficulties compared to static tasks, e.g. the importance of timing and the difficulty of reaching (and staying in) unstable regimes of the robots configuration-space. Nonetheless, robot learning for dynamic tasks has been investigated in the past including pancake flipping~\citep{kormushev2010robot}, walking on unstable surfaces~\citep{haarnoja2018soft}, and even the \bic{} task~\cite{chiappa2009using,kober2009learning}. \citet{chiappa2009using} complete the \bic{} task with a simulated robot arm using motion templates generated from human demonstration data. More directly related, \citet{kober2009learning} were able to solve the \bic{} task on a real robot, but their technique required imitation learning through kinesthetic teaching, reward shaping, a hand-designed policy class (in the form of Dynamic Movement Primitives (DMP)) and expressive features to form the state-space. In this work, we are able to learn a final policy that can solve the same task from raw image data and proprioceptive information. 

Finally, like SAC-X, our method learns a policy using an off-policy RL algorithm. We specifically utilize SVG-0 with retrace~\cite{heess2015learning,munos2016safe}. However, other off-policy RL methods such as MPO or soft-actor critic could be utilized~\cite{abdolmaleki2018maximum,haarnoja2018soft,haarnoja2018softb}. The only requirement is that the algorithm is off-policy so that the policy for one task can learn from data generated by other behaviour policies.

\section{Simultaneous Reinforcement Learning from Features and Pixels}

\subsection{Background and Notation}

In this work we consider the standard Reinforcement Learning (RL) setting as described by a Markov Decision Process (MDP). The MDP $\mathcal{M}$ consists of continuous states $s$ and actions $a$, together with an associated transition distribution $p(s_{t+1} | s_t, a_t)$ which specifies the probability of transitioning from state $s_t$ to $s_{t+1}$ under action $a_{t}$. Given a reward function $r_{\mathcal{M}}(s, a) \in \mathbb{R}$ and the discount factor $\gamma \in [0, 1)$ the goal of learning is to optimize for the parametric policy $\pi^\cM_\theta(a | s)$ with parameters $\theta$ (which specifies a distribution over actions) that maximizes the cumulative discounted reward
$
    \mathbb{E}_{\pi^\cM_\theta}\left[\sum^\infty_{t=0}\gamma^t r_{\mathcal{M}}(s_t,a_t) | \right.\allowbreak
    \left. a_t \sim \pi^\cM_\theta(\cdot | s_t),\allowbreak s_{t+1} \sim p(\cdot|s_t, a_t), s_0 \sim p(s)\right]
$, where 
$p(s)$ is the initial state distribution. This maximization problem can alternatively be defined as 
$$
J(\theta) = \mathbb{E}_{p(s)} \Big[ Q^\pi_\cM(s, a) \Big| a \sim \pi^\cM_\theta(a | s) \Big],
$$
where we define the action-value function as
$$
Q^\pi_\cM(s_t, a_t) = r_\cM(s, a) + \mathbb{E}_{\pi^\cM_\theta} \Big[ Q(s_t, \cdot) | s_{t+1} \sim p(\cdot | s_t, a_t) \Big].
$$

We are, however, not only interested in learning a single policy $\pi_\theta$ for the main task MDP $\mathcal{M}$ but, alongside, want to learn policies for a given set of auxiliary rewards $\lbrace r_{\cA_1}(s, a), \dots, r_{\cA_K}(s, a) \rbrace$. Following the notation from the Scheduled Auxiliary Control (SAC-X) framework~\cite{riedmiller2018learning} paper we thus define a set of auxiliary MDPs $\sA = \lbrace \cA_1, \cdots, \cA_K \rbrace$, that share the state-action space and the transition dynamics with $\cM$ but differ in their reward definition (e.g. MDP $\cA_i$ uses reward $r_{\cA_i}(s, a)$ respectively). We then aim to learn a policy for each such auxiliary task -- denoting them with $\pi^{A_i}_\theta(a | s)$ respectively. We refer to these per-task policies as intentions in the following -- as their goal is to entice the agent to intently explore interesting regions of the state-action space. To collect data for this learning process we randomly choose among one of the intentions in each episode and execute it on the robot. During execution a reward vector $r(s, a) = \left[ r_\cM(s, a), r_{\cA_1}(s, a), \dots, r_{\cA_K}(s, a) \right]$ is collected (with each component denoting one of the tasks). Such a random execution of intentions was referred to as SAC-U in \citep{riedmiller2018learning}. The data collected during execution is then sent to a replay buffer $\cB$ from which policies can be learned via off-policy learning -- we defer a description of this learning process to the next section, in which we extend the SAC-X setting to varying state spaces.

\subsection{Scheduled Auxiliary Control with Varying State Spaces}
\label{sec:sac-x-varying-state-spaces}

As described above the SAC-X paper introduced the concept of training simultaneously on multiple tasks, with each task having its own reward function. We now describe how this idea can be extended to a scenario in which each task is potentially associated with a different state-space. Such a learning setup is useful if one wants to learn a final policy that uses only a subset of the available observations but wants to make use of all available information at training time. In our experiments, for example, we aim to learn a policy for the \bic{} task from vision and proprioceptive information only (i.e. $\pi_\cM$ should be conditioned only on images from two cameras as well as the robot joint positions and velocities and internal state necessary to make the task Markovian). 
During training, however, we want to exploit the fact that additional state information is available (e.g. we need to be able to track the ball in the \bic{} setup to calculate a reward signal). 

The idea, then, is to simultaneously train both feature-based and vision-based intention policies -- instead of training from scratch from raw-image data only. Ideally, the feature-based intentions will learn to solve the task quickly -- due to the lower dimensionality of the state-space and the more informative features -- and then generate meaningful experience data for the vision based intentions. This can be seen as a form of curriculum learning.

Crucially, the additional feature-based intentions can be removed after training, leaving us with the ``raw-sensor'' based policy that we desired -- eliminating the need for the expensive and carefully calibrated feature detection setup at execution time.

To describe this process more formally: Let $S^\sigma$ be the set of all measurable signals at time $t$ that make up the (potentially over specified) full state $s_t$ during training, i.e.: 
$$s_t = S^\sigma(t) =\left\{o_t^0, o_t^1, \cdots, o_t^n \right\},$$
where $o_t^i$ are the observed measurements at $t$. For example, for the \bic{} task 
the measurements include: proprioceptive features consisting of joint positions and velocities, features computed via a Vicon tracking system \cite{viconweppage} (such as ball position and velocity), and finally ``raw'' sensor data (e.g. camera images). In a standard RL setup a minimal sub-set of the components of $S^\sigma$ would be used to construct the MDP state $s_t$. For example, for the \bic{} task such a sub-set could consist of proprioceptive features together with Vicon features (disregarding vision information). Instead of committing to a single state-description we consider the case in which $S^\sigma$ can be decomposed into multiple, non-overlapping subsets of observations: 

\begin{multline*}
        \sS = \left\{ S_i^\sigma, S_j^\sigma, \cdots | S_i = \left\{ o^k, \cdots\right\} \wedge S_j^\sigma = \left\{ o^l, \cdots\right\} \wedge \right. \\
        \left. S_i^\sigma \subseteq S^\sigma \wedge S_j^\sigma \subseteq S^\sigma \wedge  S_i^\sigma \cap S_j^\sigma = \varnothing \right\}.
\end{multline*}
In our experiments we use three such separate feature sets that we call $S^\sigma_\text{proprio}$ (containing robot joint position, velocities, previous actions, and additional internal robot state data), $S^\sigma_\text{features}$ (containing information about the ball position and velocity and cup position, orientation, and velocity as computed based on Vicon tracking) and $S^\sigma_\text{image}$ (the images collected from two cameras setup in the workspace), respectively.

We then associate a fixed masking or filter-vector with each of these newly constructed $S_i^\sigma$ sets given by the index vector
\begin{equation} \label{eq:state-set-filter}
    S_i^{filter} = \left[ \mathbb{I}^i(task_0),\mathbb{I}^i(task_1), \cdots, \mathbb{I}^i(task_N) \right],
\end{equation}
where the indicator function $\mathbb{I}^i(task_k)$ is defined on a per-task basis and returns 1 if this state set should be enabled for task $k$ and 0 otherwise. To simplify the presentation (but without loss of generality) we will restrict the following derivation to the concrete state-feature set $\sS = \lbrace S^\sigma_\text{proprio}, S^\sigma_\text{features}, S^\sigma_\text{image} \rbrace$ -- as mentioned above and further described in Table~\ref{tab:state-space}. 
Given these definitions, we can perform learning using the following policy evaluation and policy improvement steps. We first define two filter vectors for each task, one used for learning an action-value critic -- referred to as $S_{Q,i}^{filter}$ -- and one for optimizing the policy -- referred to as $S_{\pi,i}^{filter}$. For the rest of the paper we assume that for all tasks both filter masks enable sufficient observations to form a Markov state (i.e. they enable $S^\sigma_\text{proprio}$ together with either $S^\sigma_\text{features}$ or $S^\sigma_\text{image}$).

\paragraph{Policy Evaluation} We then construct the learned action-value functions $\hat{Q}^\pi_\cT(s_t, a_t; \phi)$ for all $\cT \in \lbrace \cM, \cA_1, \dots, \cA_K \rbrace$ with parameters $\phi$ as the following feed-forward neural network. Let \begin{equation}
\begin{aligned}
    &\hat{Q}^\pi_\cT(s_t, a_t; \phi) = f([\mathbf{e}_1 \odot \mathbf{g}_{1,:}, \mathbf{e}_2 \odot \mathbf{g}_{2,:}, \mathbf{e}_3 \odot \mathbf{g}_{3,:} ]; \phi_\cT), \\
    &\mathbf{g} = [ g_{\phi_p}(S^\sigma_\text{proprio}(t)), g_{\phi_f}(S^\sigma_\text{features}(t)), g_{\phi_i}(S^\sigma_\text{image}(t)) ]^T, \\
    &\mathbf{e} = [ \mathbb{I}^\text{Q,proprio}(\cT), \mathbb{I}^\text{Q,features}(\cT),
    \mathbb{I}^\text{Q,image}(\cT)]
\end{aligned}
\end{equation}
where $\mathbf{g}$ consists of the concatenated output of three function approximators (here feed-forward neural networks with parameters $\phi_p, \phi_f$ and $\phi_i$ respectively) applied to the state-features from each feature group, $\odot$ denotes element wise multiplication, subscripts denote element (in case of $\mathbf{e}_i$) and row-access (in case of $g_{i,:}$) respectively. $f$ denotes the final output layer and the full parameters of the network are given by $\phi = \lbrace \phi_\cM, \phi_{\cA_1}, \dots, \phi_{\cA_K}, \phi_p, \phi_f, \phi_i \rbrace$. That is, depending on whether or not a given set of features $S^\sigma_i \in \sS$ is activated for the Q-function of task $i$ the output of the corresponding $g$-network is passed forward. This way, the g-networks can be learned jointly across multiple tasks, while the output layers (given by the function $f$) specialize on a per task basis, where a task is a combination of a reward function and a state-space. Specializing outputs per task, rather than per reward, allows the network to adjust the policy to differences in the observation types such as parts of the environment that are noisy for one observation type but not another. Learning of all action-value functions is performed jointly (analogously to the procedure from \citet{riedmiller2018learning}) by regressing the retrace \citep{munos2016safe} targets: 
\begin{equation}
\begin{aligned}
  &\min_\phi L(\phi) = \sum_\cT \mathbb{E}_{\tau \sim \mathcal{B}} \Big[ \big( \hat{Q}^\pi_\cT(s, a; \phi) -
  Q^{\text{ret}} \big)^2 \Big], \text{with } \\ 
  &Q^{\text{ret}} = \sum_{j=i}^\infty \Big( \gamma^{j-i} \prod_{k=i}^j c_k \Big) \Big[ r_\cT(s_j, a_j) + \delta_Q(s_{j+1}, s_j) \Big], \\
  &\delta_Q(s_i, s_j) = \gamma \mathbb{E}_{\pi^\cT_{\theta'}(a | s)}
  [ Q^\pi_\cT(s_i, \cdot; \phi') ] - Q^\pi_\cT(s_j, a_j; \phi'), \\
  &c_k = \min\Big(1, \frac{\pi^\cT_{\theta'}(a_k | s_k, \cT)}{b(a_k | s_k)}\Big),
\end{aligned}
\end{equation}
where $b$ are the policy probabilities under which the trajectory $\tau$ was recorded and $\phi'$ denote the parameters of a target network that are copied from the current parameters $\theta$ every $1000$ optimization steps to stabilize learning \citep{mnih2015human}.
\paragraph{Policy Improvement}
Analogously to the learned action-value functions we define the policy networks as:
\begin{equation}
\begin{aligned}
    &\pi^\cT_\theta(\cdot | s_t) = \mathcal{N}(\mu_\cT, \mathbf{I}\sigma^2_i) \\
    &[\mu_\cT, \sigma^2_\cT] = f([\mathbf{e}_1 \odot \mathbf{g}_{1,:}, \mathbf{e}_2 \odot \mathbf{g}_{2,:}, \mathbf{e}_3 \odot \mathbf{g}_{3,:} ]; \theta_\cT),  \\
    &\mathbf{g} = [ g_{\theta_p}(S^\sigma_\text{proprio}(t)), g_{\theta_f}(S^\sigma_\text{features}(t)), g_{\theta_i}(S^\sigma_\text{image}(t)) ]^T, \\
    &\mathbf{e} = [ \mathbb{I}^{\pi,\text{proprio}}(\cT), \mathbb{I}^{\pi,\text{features}}(\cT),
    \mathbb{I}^{\pi,\text{image}}(\cT)],
\end{aligned}
\end{equation}
where $\mathcal{N}(\mu, \mathbf{I}\sigma^2)$ denotes the multivariate normal distribution with mean $\mu$ and diagonal covariance matrix and the parameters of the policy $\theta = \lbrace \theta_\cM, \theta_{\cA_1}, \dots, \theta_{\cA_N}, \theta_p, \theta_f, \theta_i \rbrace$ are optimized by following the reparameterized gradient of $J(\theta)$ based on the learned Q-functions:
\begin{equation}
    \nabla_\theta J(\theta) \approx \sum_\cT \nabla_\theta \mathbb{E}_{a \sim \pi^\cT_\theta(\cdot | s)}\Big[  \hat{Q}^{\pi}_\cT(s_t, a; \phi) - \alpha \log \pi^\cT_\theta(a | s_t) \Big],
\end{equation}
where the second term denotes additional entropy regularization of the policy (with weight $\alpha$). We refer to \citet{riedmiller2018learning} for details on the calculation of this gradient.

\begin{figure}[h]
    \centering
    \includegraphics[width=.75\linewidth]{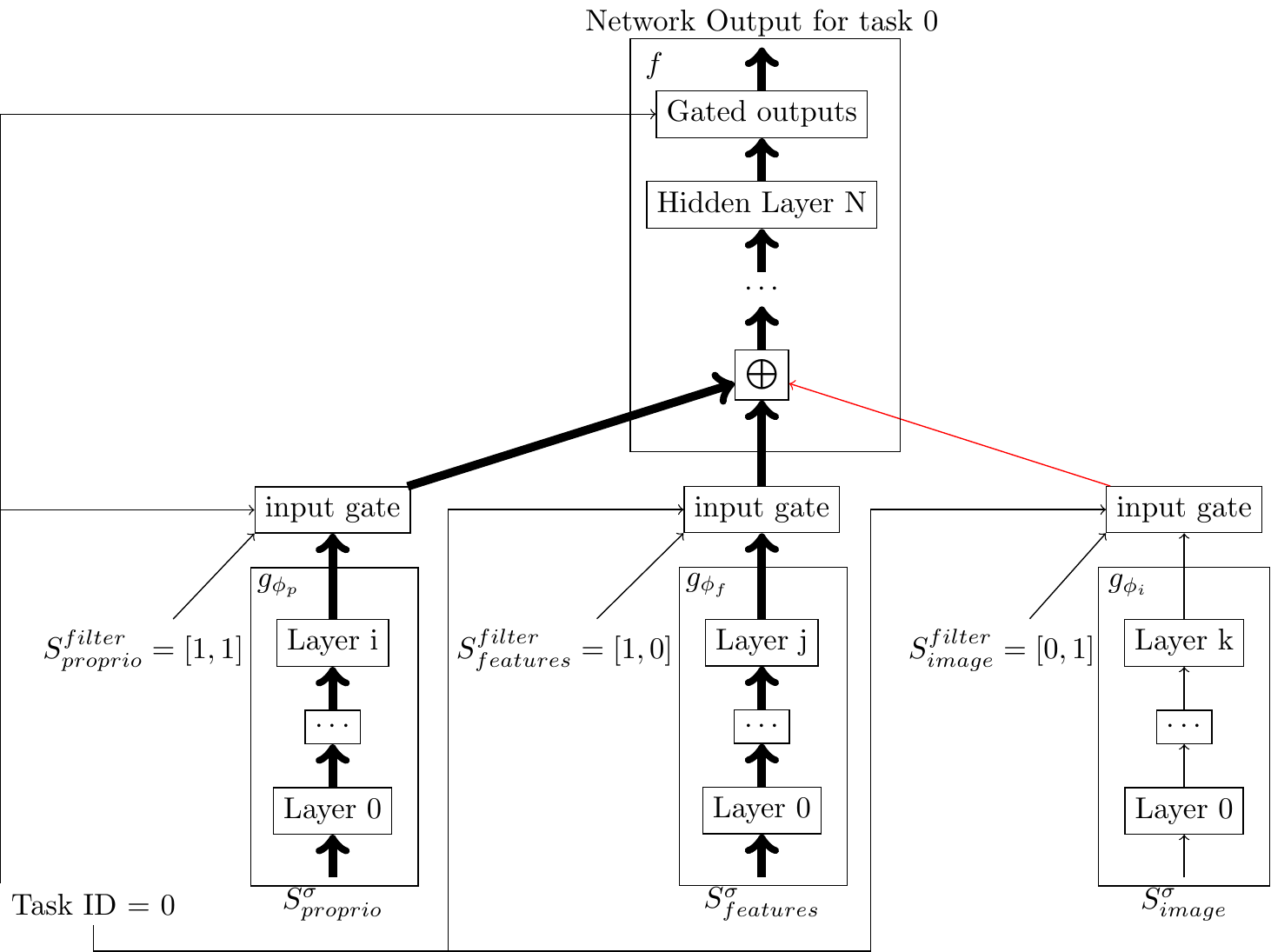}
    \caption{\label{fig:diff-state-space-network-architecture} Overall network architecture with specific example values for two tasks evaluated on task 0 and three state sets. Bold lines show which paths in the network are propagated to the output and allow gradients to flow. The red line indicates where the gate zeros the output and as a side-effect prevents gradients from flowing backwards.}
    \vspace{-.5cm}
\end{figure}

Figure~\ref{fig:diff-state-space-network-architecture} depicts the complete network architecture of a policy network for an exemplary feature selection. The input layers all use the previously described input gating architecture. The output layers use the gated output heads introduced in the SAC-X paper. In this particular example, only the proprioception and feature groups are enabled for task 0 -- and are hence allowed to flow into the hidden layers of the $f$ network -- while the image network output is zeroed out. The $f$ network used in our experiments assumes the outputs of all the $g$ functions are the same shape, and then sums the inputs element-wise. This reduces the amount of zeros flowing into the $f$ network layers, but requires the assumption that each of the state-spaces can match the expected internal representation of each other.

\subsection{Asymmetric Actor-Critic through Different State Spaces}
\label{sec:asym-actor-critic-diff-state-spaces}

The idea of using different observations for different tasks that we described above has similarities to the asymmetric actor-critic technique from~\citet{pinto2017asymmetric} that are worth expanding on. In asymmetric actor-critic training, the state-space of the critic (in our case the Q-function network) is enriched with additional information (which may not be available at test time). This additional information (e.g. extracted, hand-designed features) can be used by the critic to simplify the problem of credit-assignment (identifying the cause of reward for a given state-action pair). Once a policy has finished training the critic is no longer needed, and can be discarded -- akin to how auxiliary tasks can be discarded after training with our method. 
When viewed from this perspective, we can interpret our method as an extension of asymmetric actor-critic to a multi-task setting (with varying state-definitions across task-policies).  

Furthermore, we can observe that the ideas of auxiliary tasks with state-based features (to aid fast learning from raw images) and asymmetric learning in actor and critic as complimentary. This opens up further possibilities in our experiments: in addition to learning both feature and vision based policies we can restrict the critic state-filters ($S^{\text{filter}}_{Q,i}$) to provide only feature observations for all tasks, potentially leading to an additional speed-up in learning time (we refer to this as the asymmetric actor-critic setting below).

\section{\bic{} Task Setup}
The following section describes the experimental setup for both simulation and the real-robot. Table~\ref{tab:state-space} shows the observations contained in the previously defined state-groups (i.e. $S^\sigma_\text{proprio}$, $S^\sigma_\text{features}$, $S^\sigma_\text{image}$), as well as the source of the observation and its shape.

\begin{table}[h]
    \centering
    \resizebox{.85\linewidth}{!}{
    \begin{tabular}{rrrc}\toprule
    \textbf{State-group} & \textbf{Data Source} & \textbf{Observation} & \textbf{Shape}\\\midrule
    $S^\sigma_\text{features}$\\\midrule
    & Vicon\\\cmidrule{2-4}
    & & Cup Position & 3\\
    & & Cup Orientation (Quaternation) & 4\\
    & & Ball Position & 3\\
    & Finite Differences\\\cmidrule{2-4}
    & & Cup Linear Velocity & 3\\
    & & Cup Angular Velocity (Euler Angles) & 3\\
    & & Ball Linear Velocity & 3\\
    $S^\sigma_\text{proprio}$\\\midrule
    & Sawyer\\\cmidrule{2-4}
    & & Joint Position & 7\\
    & & Joint Velocities & 7 \\
    & Task \\\cmidrule{2-4}
    & & Previous Action & 4 \\
    & & Action Filter State & 4\\
    $S^\sigma_\text{images}$\\ \midrule
    & & Stacked Color Front Image & $84 \times 84 \times 3 \times 3$\\
    & & Stacked Color Side Image & $84 \times 84 \times 3 \times 3$\\\bottomrule
    \end{tabular}}
    \caption{\label{tab:state-space} State group definitions with observation sources and shapes.}
    \vspace{-.25cm}
\end{table}

We use a Sawyer robot arm from Rethink Robotics~\footnote{http://mfg.rethinkrobotics.com/intera/Sawyer\_Hardware}, which has 7 degrees of freedom~(DoF). The agent controls velocities of 4 out of 7 DoFs (joints J0, J1, J5 and J6) -- which are sufficient to solve the task. The unused DoFs are commanded to remain at a fixed position throughout the episode.

On the real system, the position and orientation features of the ball and cup are determined by an external Vicon motion capture system. IR reflective markers are placed on the cup and ball. The Vicon system tracks the position and orientation of the marker clusters. The position and orientation estimates are accurate to sub-millimeter precision, have low latency and low noise.

\begin{figure}
    \centering
    \includegraphics[width=.7\linewidth]{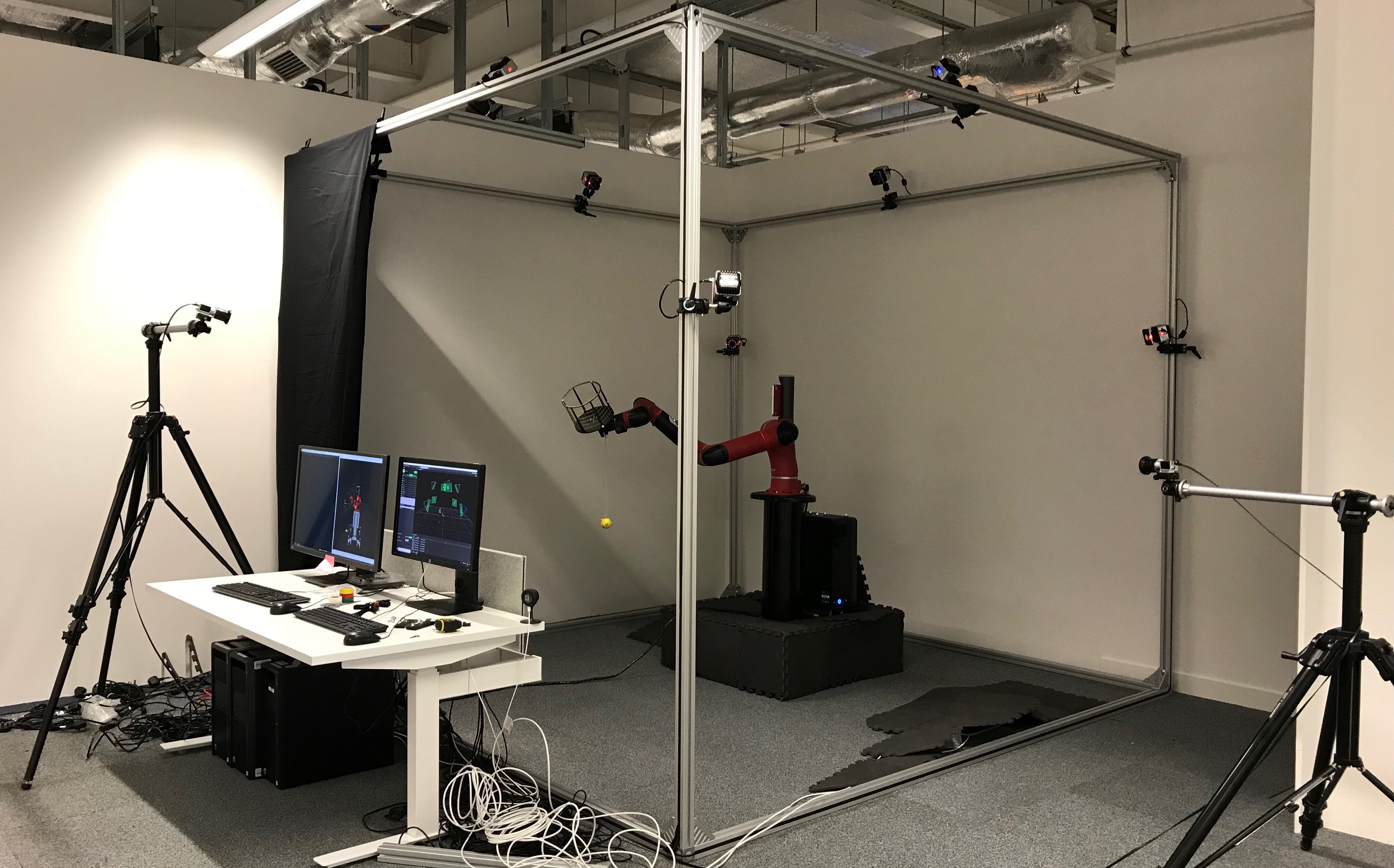}
    \caption{\label{fig:real-robot-setup}Picture of the real robot setup.}
    \vspace{-.25cm}
\end{figure}

\begin{figure}
    \centering
    \includegraphics[width=.65\linewidth]{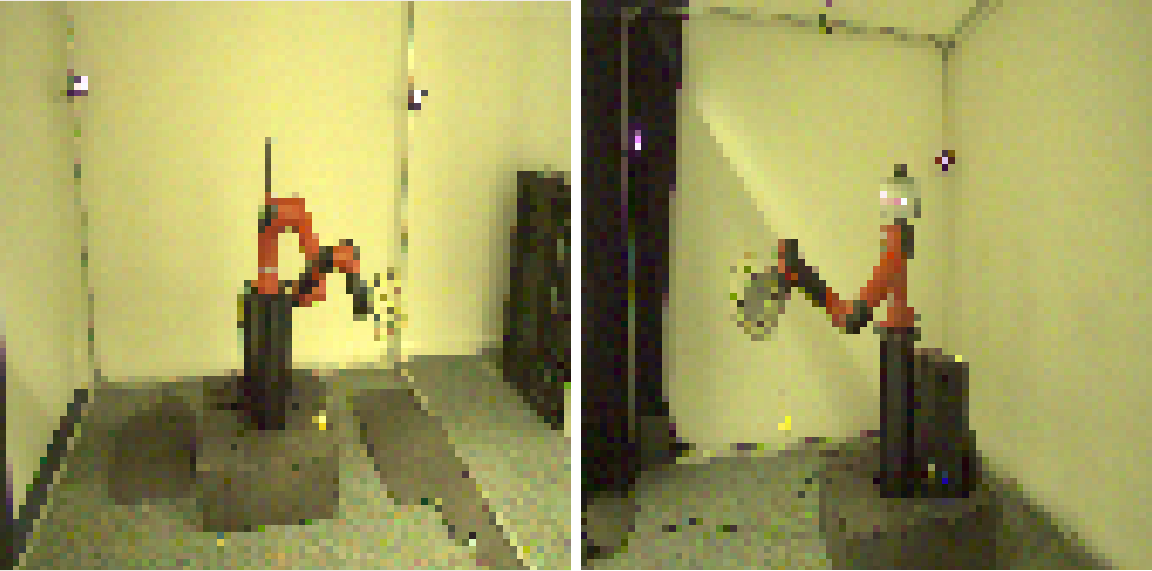}
    \caption{\label{fig:real-robot-cameras}Agent's perspective: down-sampled front camera (left) and down-sampled side camera (right) with ($84\times84$) pixels.}
    \vspace{-.25cm}
\end{figure}

We also include two external cameras roughly positioned at orthogonal views (see Figure~\ref{fig:real-robot-setup}). RGB frames of $1920\times1080$ pixels are captured at 20 frames per second, and down-sampled via linear interpolation to $84\times84$ pixels -- and subsequently fed to the neural networks. Figure~\ref{fig:real-robot-cameras} shows pictures of the real camera images seen by the robot during training. We deliberately make no effort to further crop the images, tune the white balance or correct for lens distortions -- relying on the adaptability of our approach instead. To enable estimation of ball velocities, when learning from raw images, we feed a history of the last three images captured by the cameras into the neural networks.

Data from proprioception (100Hz), Vicon ($\sim$100Hz) and cameras (20Hz) is received asynchronously. State observations are sampled from this data at 20Hz with no buffering (\textit{i.e.}, most recent data is used). The agent controls the robot synchronously with the observation sampling at 20 Hz. The control rate is chosen to allow for the robot to dynamically react to the current ball position and velocity, while not being so high-rate as to overwhelm the learning algorithm.

\subsection{Experimental setup}

The ball has a diameter of 5~cm and is made of foam to prevent damage to the robot while swinging. A Kevlar string of 40~cm was used to attach the ball to a ball bearing fixed at the robot's wrist. The bearing is in line with the robot's wrist axis of rotation and helps prevent winding of the string while swinging the ball -- when the ball goes around the wrist the string rotates freely via the bearing.

The cup is 3D printed, and a woven net is attached to it to ensure visibility of the ball even when inside the cup. The cup is  attached to the wrist right behind the bearing. We perform the experiment with two different cup sizes: a large cup with 20~cm diameter and 16~cm height and a small cup with 13~cm diameter and 17~cm height. Experiments are performed with the larger cup size unless otherwise stated.

\subsection{Robot Safety}

When training on the real robot, the safety of the robot and its environment must be considered. An agent sending random, large, velocity actions to the robot may cause it to hit objects or itself. Additionally, quickly switching between minimum and maximum joint velocity commands may incur wear to the robot's motors. We hence utilize an action filter as well as a number of hard-coded safety checks to ensure safe operation.

In more detail: The Sawyer robot has built in self-collision avoidance, which is enabled for the task. To prevent collisions with the workspace we simply limit the joint positions to stay within a safe region. 
Additionally, the limits help keep the real robot, cup and ball within the field of view of the Vicon system and cameras.

To prevent high frequency oscillations in the velocities we pass the network output through a low-pass first order linear filter with cutoff frequency of 0.5~Hz -- which was chosen qualitatively so that the robot is able to build up large velocities before hitting the joint limits but prevents switches between minimum and maximum velocities. Such a filter has an internal state which depends on the history of policy actions. If unobserved, this would make the learning problem non-Markovian. We hence, provide the agent with the internal state of the action filter as part of the proprioceptive observations.

\subsection{Episode Resets}

Training is episodic, with a finite horizon of 500 steps per episode, resulting in a total wall clock time of 25 seconds per episode. We randomly switch between intention policies every 100 steps (dividing the episodes into 5 subsets) -- this is done to increase diversity in the starting positions for each policy. In simulation, episodic training is easy, as the simulator can simply be reinitialized before each episode. In the real world, resets for the \bic{} task can be tricky -- due to the string wrapping and tangling around the robot and cup. In order to minimize human intervention during training, the robot uses a simple hand-coded reset policy that allows for mostly unattended training. This procedure simply commands the robot joints to drive to positions from a set of predefined positions at random (that were chosen to maximize the number of twists the arm does in order to untangle the string). For more details on the reset behaviour we refer to the supplementary material.

\subsection{Simulation}

\begin{figure}
    \centering
    \includegraphics[width=.75\linewidth]{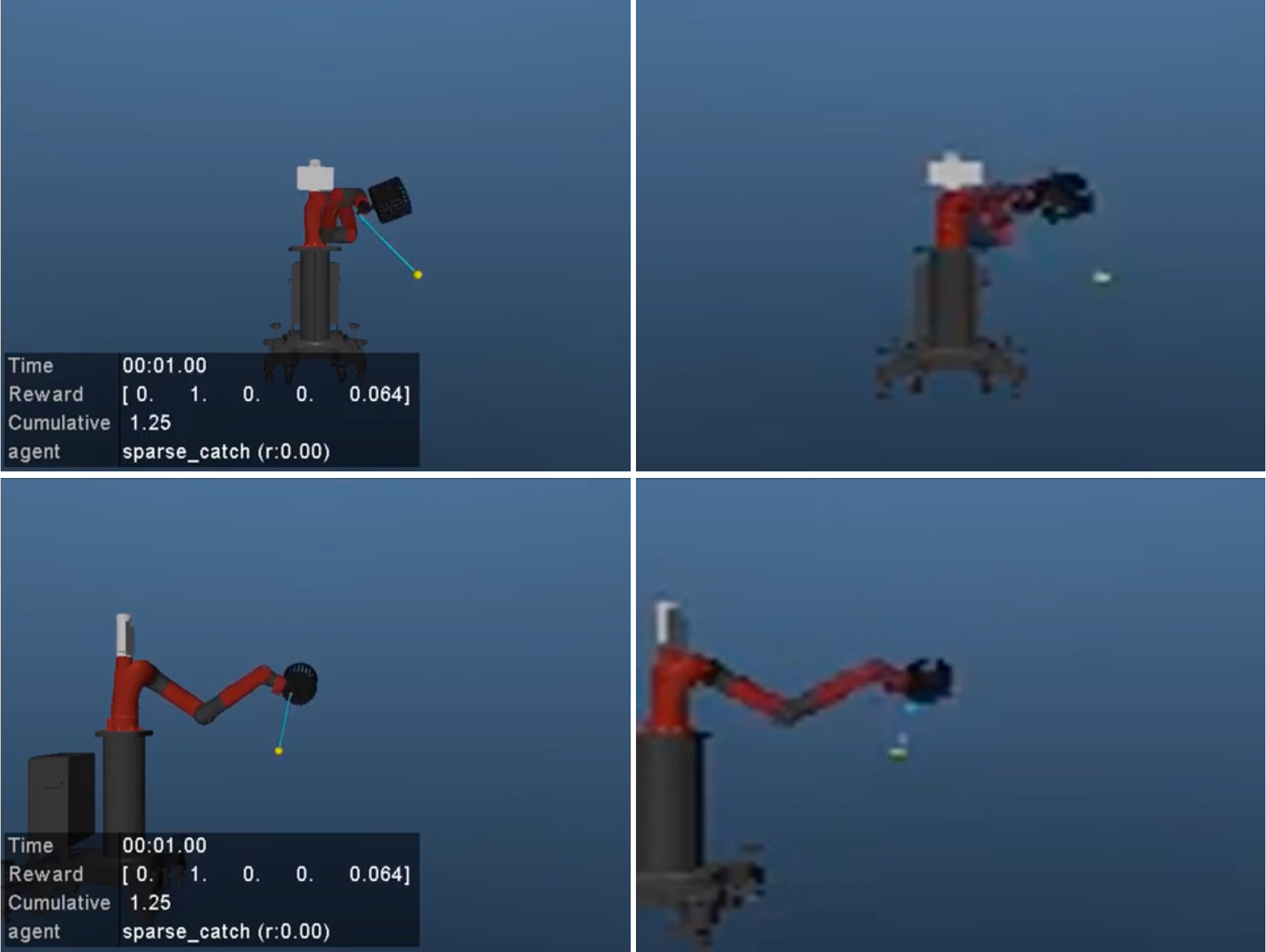}
    \caption{\label{fig:simulation-screenshots} Example of the simulation (left column) along with what the agent sees when using the images (right column).}
    \vspace{-.25cm}
\end{figure}

We use the MuJoCo physics engine to perform simulation experiments. We use a model of the robot using kinematic and dynamic parameters provided by the manufacturer. Joint torques and maximum velocity values are also taken from a data-sheet. We do not attempt to create a highly accurate simulation (i.e. via system identification), as we mainly use the simulation to gauge asymptotic learning behaviour. When training in simulation, observations come directly computed from the simulation state and hence are noise free -- e.g. images are rendered with a plain background and do not contain artifacts that real-cameras suffer from such as distortion or motion blur. 
Figure~\ref{fig:simulation-screenshots} shows the simulation setup as well as samples of the raw camera images provided to the agent.

\section{Experiments}

\subsection{Task Descriptions}

Table~\ref{tab:task-reward-functions} shows the reward functions used in the various simulation and real-robot experiments, here the ball position $(x, y, z)$ is always w.r.t.\ the cup coordinate frame. Note that $r_5$ is the main sparse reward function for the \bic{} task.

\begin{figure}
    \centering
    \includegraphics[width=.9\linewidth]{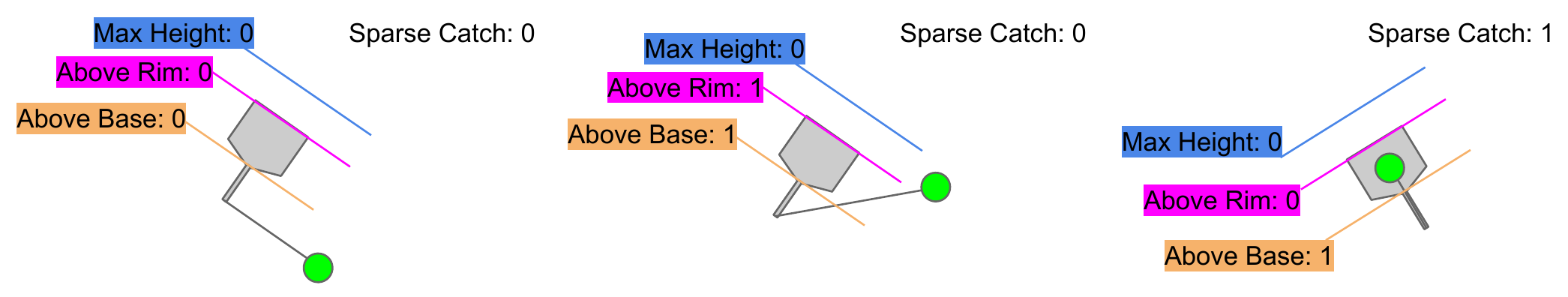}
    \caption{\label{fig:cup-components} Parts of the cup referred to in the task reward function definitions.}
    \vspace{-.25cm}
\end{figure}

Figure~\ref{fig:cup-components} illustrates the different cup reference frames and positions used for the reward computation, such as the base, rim, etc.. The formula for $r_6$ is defined according to Equation~\ref{eq:shaped_cup_base}, which was tuned to return $\left[0, 1\right]$ based on the minimum and maximum ball height, respectively. The reward shaping of $r_7$ uses a 2D Gaussian with $\mu = 0$ (cup base center) and $\sigma = 0.09$ for both X and Y axes, as per Equation \ref{eq:shaped_swing-up}. In this case, $\sigma$ is defined so that values within the interval $\left[-2\sigma, 2\sigma\right]$ would coincide with the cup diameter and return zero when the ball is below the cup base;

\begin{align}
    r_{6} &= \frac{1 + \tanh(7.5 z)}{2},\label{eq:shaped_cup_base}\\
    r_{7} &= \frac{1}{2\pi\sigma^2} e^{\frac{-(x^2 + y^2)}{2\sigma^2}}\label{eq:shaped_swing-up}.
\end{align}

As discussed in section~\ref{sec:sac-x-varying-state-spaces}, we use three state-groups: $S^\sigma_\text{proprio}$, $S^\sigma_\text{features}$, and $S^\sigma_\text{images}$. Table~\ref{tab:state-space} shows which observations belong to which state-groups.

We use two main state-spaces for the different training task which we refer to as: feature state-space and pixel state-space. Feature state-space tasks have both $S^\sigma_\text{proprio}$ and $S^\sigma_\text{features}$ enabled. Pixel state-space tasks have both $S^\sigma_\text{proprio}$ and $S^\sigma_\text{image}$ enabled.

Tasks are defined by a combination of state space type and reward function. We refer to a specific combination by the reward id followed by either ``F'' for feature state space or ``P'' for pixel state space. For example, 1F refers to a task with reward function 1 and a feature state-space.

\begin{table}[h]
    \centering
    \resizebox{.85\linewidth}{!}{
    \begin{tabular}{|c|l|p{3cm}|}\hline
        \textbf{Reward ID} & \textbf{Reward Name} & \textbf{Reward Function} \\\hline
        1 & Ball Above Cup Base & +1 if ball height is above the base of cup in the cup frame \\\hline
        2 & Ball Above Cup Rim & +1 if ball height is above rim of cup in cup frame \\\hline
        3 & Ball Near Max Height & +1 if ball height is near the maximum possible height above cup in cup frame\\\hline
        4 & Ball Near Opening & Shaped distance of ball to center of cup opening \\\hline
        5 & Sparse Catch & +1 if ball in cup \\\hline
        6 & Shaped Ball Above Cup Base & See equation \ref{eq:shaped_cup_base} \\\hline
        7 & Shaped Swing up &See equation \ref{eq:shaped_swing-up}\\\hline
        8 & Do nothing (distractor task) & Negative reward proportional to joint velocities\\\hline
    \end{tabular}}
    \caption{\label{tab:task-reward-functions} Reward functions used for the different tasks. Reward 5 is the main reward for the actual \bic{} task.}
    \vspace{-.25cm}
\end{table}

\subsection{Learning with Different State-Spaces in Simulation}

To measure the effectiveness of our proposed approach we first perform an ablation study in simulation -- so that training runs can be parallelized and meaningful statistics can be easily gathered. Our evaluation protocol is as follows: after each episode of training we freeze the weights of the main task policy and run an evaluation episode for the sparse catch task (using the appropriate state space for this task) and report the cumulative reward.

\begin{figure}
    \centering
    \includegraphics[width=.8\linewidth]{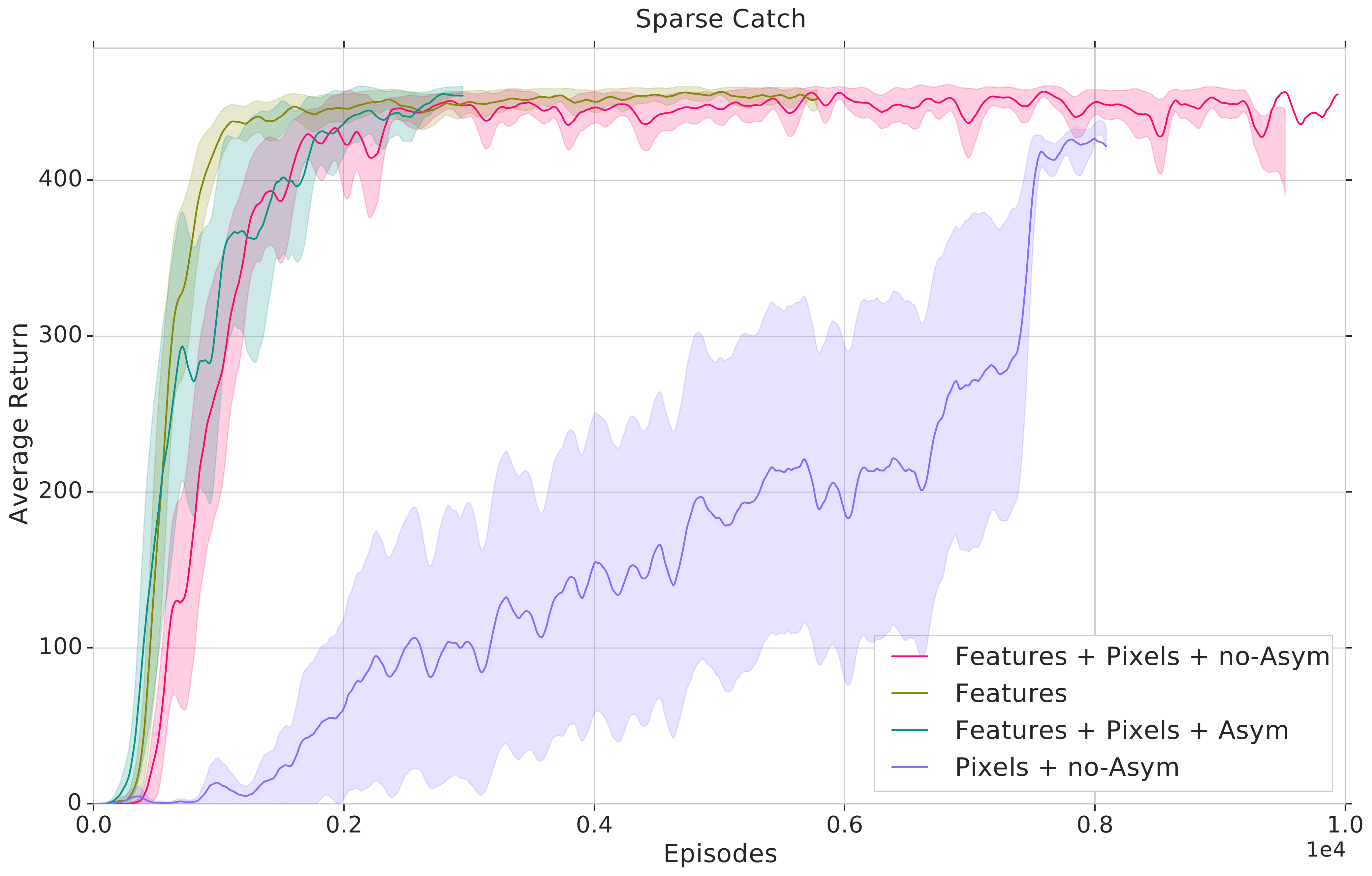}
    \caption{\label{fig:simulation-diff-state-vs-baseline} Comparison of simultaneously learning in different state spaces vs baselines in simulation. Solid lines show mean performance across 10 independent runs.}
    \vspace{-.25cm}
\end{figure}

Figure~\ref{fig:simulation-diff-state-vs-baseline} shows the results of the ablation experiment. Each curve in the plot depicts the mean of 10 independent runs of the average reward of the evaluated task. The yellow curve shows the evaluation of task 5F when trained with tasks 1F, 2F, 3F, 4F, and 5F. As hypothesized earlier, training based on feature-only observations converges fastest -- presumably because it is easier to learn within the feature state-space given its low dimensionality and expressiveness. With an episode time of 20 seconds, learning succeeds in time equivalent to approximately 5.5 hours of real-time training. The purple curve shows the evaluation of task 5P when trained with tasks 1P, 2P, 3P, 4P, and 5P. Using only image based state-space definitions results in approximately 8 times slower convergence of the main task policy. For comparison, this corresponds to an estimated 1.6 days of continuous training time on a real robotic system. In practice, the total time to train would likely be even higher as the episode resets takes time in the real world (and our simulation is noise-free).

Finally, the red and green curves show the results of our method. The red curve depicts the evaluation of task 5P when training is performed with tasks 1F, 2F, 3F, 4F, 5F, 1P, 2P, 3P, 4P and 5P, with $\forall i, S^{filter}_{Q,i} = S^{filter}_{\pi,i}$. We can observe that this results in a significant speed-up in learning time, when compared to training from image data alone -- with training time being slowed down only by a factor of 2, compared to training from features. The green curve shows the evaluation of task 5P when trained with tasks 1F, 2F, 3F, 4F, 5F, 1P, 2P, 3P, 4P and 5P, but this time using the asymmetric actor-critic technique (i.e. $S^\sigma_\text{features}$ enabled in all critic filter lists and $S^\sigma_\text{image}$ disabled in all critic filter lists). We see that there is a slight improvement in the early training regime, likely due to faster convergence of the critic network. This shows that faster learning of a good value estimate can lead to additional speed-ups.

\subsection{Learning from Features on a Real Robot}

The simulation results from the previous section suggest that our approach can result in a significant speed-up when compared to learning from raw images only. However, there clearly exist differences between simulation and the real-robot due to modeling inaccuracies: Firstly, the simulated parameters of the robot (e.g. actuator torques, joint frictions, etc.) likely not closely matching the real robot. Secondly, the sensors in simulation are perfect with no noise and no missing observations unlike their real world counterparts. To gauge whether learning on the real robot performs similarly we hence first verified that we can learn a feature-based policy on the real robot.

\begin{figure}
    \centering
    \includegraphics[width=.8\linewidth]{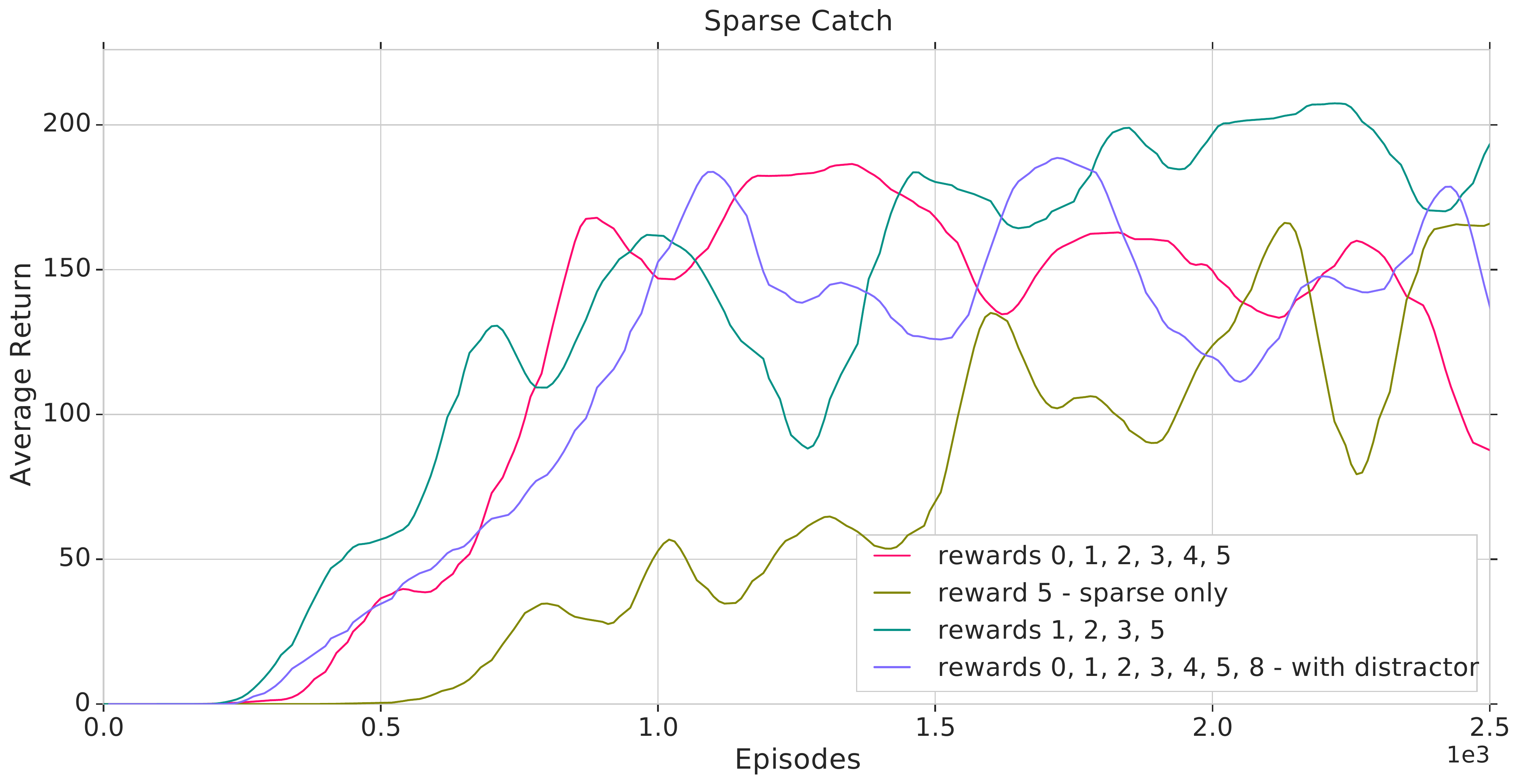}
    \caption{\label{fig:real-robot-features-learning-curves} Real robot learning curve of task 5F for runs learning the \bic{} task with combinations of tasks 1F, 2F, 3F, 4F, 5F and 8F.}
    \vspace{-.25cm}
\end{figure}

Fort this we used different combinations of tasks 1F, 2F, 3F, 4F, 5F and 8F together with the larger cup. Figure~\ref{fig:real-robot-features-learning-curves} shows the resulting average reward of task 5F as training proceeds on the real robot. Note that only a single robot is used. Training takes place with minimal human supervision (a human operator only intervenes to reset the environment when the robot is unable to reset itself after multiple tries).

The yellow curve shows the performance of the policy for task 5F when only task 5F is used for training. This demonstrates that it is possible to learn \bic{} on the real robot with just the sparse reward function. However, the other curves in the plot show that training speed can be improved if good auxiliary tasks are included during training. This finding is in agreement with the prior SAC-X work~\citep{riedmiller2018learning}. More interestingly, the purple curve shows the result of training with a purposefully useless ``distractor'' auxiliary task (8F), that is added to the auxiliary tasks 1F, 2F, 3F. Even with this distractor task, the robot learns a good sparse catch policy from features -- using approximately the same training time as when only the useful tasks are included. This suggests that, while it may be difficult to find a minimal subset of auxiliary tasks which speed-up training, it may be safe to just add any additional auxiliary tasks one thinks to be useful, with minimal effect on the training time and the final policy performance. 

Qualitatively, the learned policies exhibit good behavior. The swing-up is smooth and the robot recovers from failed catches. With a brief evaluation of 20 runs, each trial running for 10 seconds, we measured 100\% catch rate. The shortest catch time being 2 seconds. 

We repeated the same experiments with the smaller cup, to increase the difficulty of the task and assess achievable precision and control. There was a slight slow-down in learning and a small drop in catch rate to 80\%, still with a shortest time to catch of 2 seconds.

\subsection{Learning with Different State-Spaces on a Real Robot}

Finally, we applied our full method to obtain an image-based policy for the \bic{} task. Due to the long experiment times on a real robot, we did not re-run all baselines from Figure~\ref{fig:simulation-diff-state-vs-baseline}. Instead we opted to only use the approach that worked best in simulation: learning with different state-spaces combined with asymmetric actor-critic. We initially used tasks 1F, 2F, 3F, 4F, 5F, 1P, 2P, 3P, 4P, and 5P as in the simulation experiments. While all of the task policies showed some improvement in this experiment learning was slower than in simulation. To minimize experimentation time we hence decided to change the auxiliary rewards into shaped rewards, with the intention of speeding-up training time for these auxiliary tasks in order to more quickly generate good data for the sparse catch policy. A description of the changed reward structure can be found in the supplementary material.

\begin{figure}[ht]
    \centering
    \includegraphics[width=.7\linewidth]{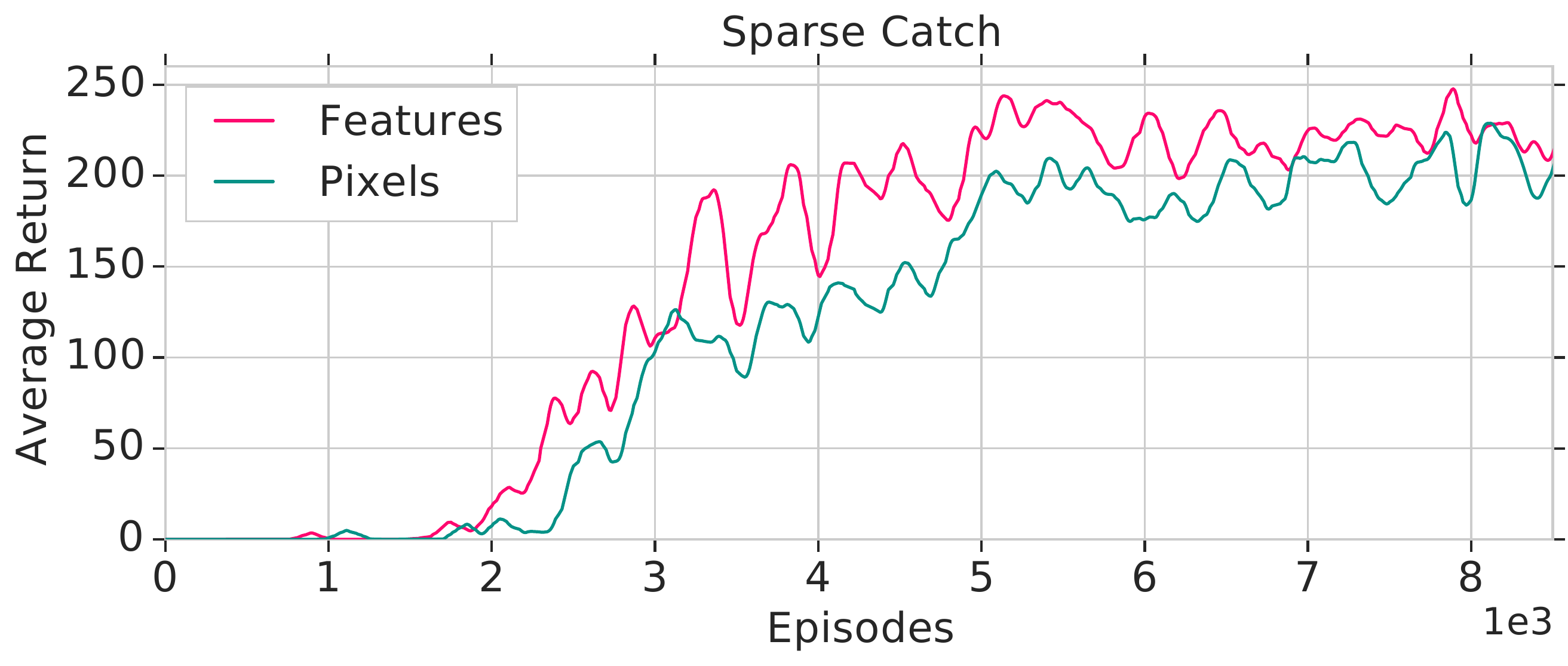}
    \caption{\label{fig:real-robot-diff-state-space-features} Learning curve for task 5F and 5P. Training was done with tasks 5F, 6F, 7F, 5P, 6P, and 7P. The asymmetric actor-critic technique was used.}
\end{figure}

Figure~\ref{fig:real-robot-diff-state-space-features} shows the learning curve for task 5F and 5P when trained with 5F, 6F, 7F, 5P, 6P, and 7P. The critic filter vectors enabled only $S^\sigma_\text{proprio}$ and $S^\sigma_\text{features}$ for all tasks. We can observe that the policies for both state-spaces converge to their final performance after around 5,000 episodes. Thus, the training time is longer than in the equivalent simulation experiment (c.f. Figure~\ref{fig:simulation-diff-state-vs-baseline}). This increase in training time is likely due to the differences in the quality of the simulated \textit{vs} real-world camera images. I.e. the simulation images are noise-free, have no motion blur and are consistently lit with proper colors. Furthermore, the robot joint sensors have noise and delays which are not modelled in simulation. Despite the inherent challenges of the real world, we are still able to learn a sparse catch policy (operating on image and proprioception data only) from scratch in about 28 hours -- ignoring time used for episode resets. This amounts to using about twice as many episodes as in our successful simulation experiments.

We conducted an evaluation of the final policy saved after 8,360 episodes. During the evaluation the exploration noise was set to zero. We ran a total of 300 episodes, achieving $100\%$ catch rate (catch in all 300 episodes). Results of this evaluation (total reward and catch time) are presented in Table~\ref{tab:stats-real-pixels}. 
\begin{table}[h]
    \centering
\resizebox{.7\linewidth}{!}{
    \begin{tabular}{rlll}\toprule
     & \textbf{Mean} & \textbf{Min.} & \textbf{Max.} \\\midrule
    Catch time (seconds) & 2.39 & 1.85 & 11.80 \\
    Total reward (maximum of 500) & 409.69 & 14.0 & 462.0 \\\bottomrule
    \end{tabular}}
    \caption{\label{tab:stats-real-pixels} Evaluation of pixel sparse catch policy over 300 episodes.}
    \vspace{-.25cm}
\end{table}

\section{Conclusion}

In this work we introduced an extension to the Scheduled Auxiliary Control (SAC-X) framework which allows simultaneously learning of tasks that differ not only in their reward functions but also in their state-spaces. We show that learning speed can be improved by simultaneously training a policy in a low-dimensional feature-space as well as a high-dimensional raw-sensor space. On the experimental side, we demonstrated that our method is capable of learning dynamic real robot policies from scratch. To this end, we learned the \bic{} task, from scratch, using a single robot. Unlike previous work we can learn this task without any imitation learning, or restrictions on the policy class. Once learned, our policy can operate using raw camera images and proprioceptive features only.

\bibliographystyle{plainnat}
\bibliography{references}

\clearpage


\appendices
\section{Network Architecture and Hyperparameters}
In this section we outline the details on the hyper-parameters used for our algorithm and baselines. Each intention policy is given by a Gaussian distribution with a diagonal covariance matrix, i.e. $\pi(\vec a| \vec s,\theta) = \mathcal{N}\left(\mu , \vec
  \Sigma \right)
$

The neural network outputs the mean $\mu=\mu(s)$ and diagonal Cholesky factors $A=A(s)$, such that $\Sigma = AA^T$. The diagonal factor $A$ has positive diagonal elements enforced by the softplus transform $A_{ii} \leftarrow \log(1 + \exp(A_{ii}))$ to enforce positive definiteness of the diagonal covariance matrix.

The general network architecture we use is described in Table \ref{t:hypers}. The image inputs are first processed by two convolutional layers followed by a fully-connected layer (see state-group heads size) followed by layer normalization. The other input modalities (features and proprioception) go through a fully connected layer (see state-group heads size), followed by layer normalization. The output of each of these three network blocks are then multiplied with 1 or 0 (depending on $S^\text{filter}$). Because all the input layers output the same shape, and we assume that the different state-groups can extract the same internal hidden representation, we sum the state-group output layers elementwise. This reduces the number of dead inputs and number of parameters needed in the first shared hidden layer. This summed output is passed to a set of fully-connected, shared hidden layers (shared layer sizes in the table). The shared hidden layer output is passed to the final output layer. In the actor, the output size is the number of means and diagonal Cholesky factors. In the critic, the output size is 1, corresponding to the Q-value of the state-action pair.

\begin{table}[ht]
\begin{center}
 \begin{tabular}{c||c} 
 Hyperparameters & SAC-X \\
 \hline
 2D Conv layer features (layer 1/ layer 2) & $16, 16$\\
 2D Conv shapes (layer 1/ layer 2) & $4\times4$, $3\times3$\\
 2D Conv strides (layer 1/ layer 2) & $2\times2, 2\times2$\\
 Actor net shared layer sizes & 200, 200\\
 Actor net state-group heads size & 100\\
 Critic net shared layer sizes & 400, 400\\
 Critic net state-group heads size & 200\\
 Discount factor ($\gamma$) & 0.99 \\
 Adam learning rate & 0.0001 \\
 Replay buffer size & 1,000,000 \\
 Target network update period & 1,000\\
 Batch size & 32\\
 Maximum transition use & 2,500\\
 Activation function & elu\\
 Tanh on networks input & No\\
 Tanh on output of layer norm & Yes\\
 Tanh on Gaussian mean & Yes \\
 Min variance & $10^{-2}$\\
 Max variance & 1.0 
\end{tabular}
\end{center}
\caption{Hyper parameters for SAC-X}
\label{t:hypers}
\end{table}

\section{Real World Episode Reset Procedure}
As described in the main paper we use a hand-coded reset strategy at the beginning of each episode (if the string was detected to be tangled around the arm). This procedure works as follows. Before starting the episode the robot arm is driven to a preset starting position. If the string is untangled and the ball is hanging freely, then the ball will be in the expected starting area. If the ball is not detected in this area, then the robot begins its untangle procedure. Ideally, we would know the state of the string and be able to either plan or learn a sequence of actions to untangle the string. However, the string is unobservable in features and given the low-resolution of the images, largely unobservable to the cameras. Instead, to untangle, the robot picks a random sequence of positions from a pre-set list. This list was chosen to maximize the number of twists and flips the arm does in order to maximize the chance that the string is unwrapped. After the sequence is completed the robot returns to the check position. If the ball is untangled then the next episode can begin. If the ball is still not in the expected area the entire procedure is repeated until a successful untangle.

\section{Robot workspace}
\begin{table}[t!]
\begin{center}
 \begin{tabular}{c||c c c c} 
  & J0 & J1 & J5 & J6 \\
 \hline
 min.\ position (rad) & -0.4 & 0.3 & 0.5 & 2.6 \\
 max.\ position (rad) & 0.4 & 0.8 & 1.34 & 4.0 \\
 min.\ velocity (rad/s) & -2.0 & -2.0 & -2.0 & -2.0 \\
 max.\ velocity (rad/s) & 2.0 & 2.0 & 2.0 & 2.0
\end{tabular}
\end{center}
\caption{Joint limits imposed during the experiments}
\label{tab:joint-limits}
\end{table}

\subsection{Joint limits}
The Sawyer robot arm has a very large workspace\footnote{\url{http://mfg.rethinkrobotics.com/intera/Sawyer_Hardware}} and can reach joint configurations which are not required to solve the \bic{} task or are outside of the motion capture system volume. In order to reduce the robot workspace, throughout all experiments we fixed J2, J3 and J4, and constrained position and velocity limits of the other joints, as detailed in Table~\ref{tab:joint-limits}.

\begin{table}[h!]
\begin{center}
 \begin{tabular}{c||c c c c c c c} 
  & J0 & J1 & J2 & J3 & J4 & J5 & J6 \\
 \hline
 position 1 & 0 & -2.07 & 0.0 & 0.23 & 0.0 & 0.96 & 0.0 \\
 position 2 & 0 & -0.22 & -2.23 & 0.3 & 0.0 & 1.18 & 3.0 \\
 position 3 & 0 & -1.65 & 3.0 & -0.23 & 0.0 & 0.0 & 3.22 \\
 position 4 & 0 & -0.59 & 3.0 & -0.23 & 0.0 & 0.0 & 3.22 \\
\end{tabular}
\end{center}
\caption{Pre-set positions used during untangling}
\label{tab:untangling}
\end{table}
\subsection{Starting position}
At the beginning of every episode, the robot joints J0-J6 are commanded to $[0.0, 0.5, 0.0, -1.22, 0.0, 0.68, 3.3]$.

\subsection{Untangling positions}
The pre-set list of untangling positions is detailed in Table~\ref{tab:untangling}.

\section{Changes to Experiment Reward Functions}

As mentioned in the main text, the experiment in section V-D used a set of modified reward functions. The initial real robot training used tasks 1F, 2F, 3F, 4F, 5F, 1P, 2P, 3P, 4P, and 5P. As discussed, these rewards were switched to shaped rewards to speed-up training time.

The actual tasks used in section V-D are 5F, 6F, 7F, 5P, 6P, and 7P. Tasks 1F, 2F and 3F have been replaced by shaped reward 6F. Task 4F was replaced by shaped reward 7F. Similarly, tasks 1P, 2P and 3P are replaced by shaped reward 6p. Task 4P was replaced by shaped reward 7P.

\end{document}